\documentclass[11pt, a4paper, copyright, goog]{google}

\usepackage[numbers, sort&compress]{natbib}

\usepackage{ifthen}
\usepackage{makecell}
\usepackage{nicefrac}
\usepackage[most]{tcolorbox}
\usepackage{ragged2e}
\usepackage{setspace}
\usepackage{subcaption}
\usepackage{float}
\usepackage{listings}

\newboolean{ANONYMIZE}
\setboolean{ANONYMIZE}{false}

\usepackage{titlesec}
\titlespacing*{\section}{0pc}{2.76ex plus 3.7pt minus 2.8pt}{4.6pt}
\titlespacing*{\subsection}{0pc}{2.3ex plus 2.76pt minus 1.84pt}{1.84pt}
\titlespacing*{\subsubsection}{0pc}{1.84ex plus 2.3pt minus 1.4pt}{1.84pt}

\paperurl{}
\renewcommand{\today}{}

\uselogo{}

\title{Agentic Framework for Deep Learning workload migration via In-Context Learning}
\correspondingauthor{sethusankaran@google.com}
\author[1]{Qiyue Liang}
\author[1]{Steven Ingram}
\author[2]{George Vanica}
\author[2]{Andi Gavrilescu}
\author[1]{Newfel Harrat}
\author[3]{Hassan Sipra}
\author[4]{Sethuraman Sankaran}

\affil[1]{Google, Sunnyvale, CA}
\affil[2]{Google, Kirkland, WA}
\affil[3]{Google, Mountain View, CA}
\affil[4]{Google, New York, NY} 

\begin{abstract}
\vspace{-1.2em}
{\normalfont\mdseries\noindent Code: \url{https://github.com/AI-Hypercomputer/accelerator-agents/tree/main/MaxCode}}\\[1.2em]
Translating deep learning models from PyTorch's flexible, object-oriented design to JAX's functional, stateless setup is usually a manual and error-prone task. Automated migration is challenging because Large Language Models (LLMs) struggle with strict and dynamic API alignment and are prone to mistakes for exacting operations. We propose a fully autonomous system that combines In-Context Learning (ICL) with oracle-driven self-debugging. First, we curated an ICL context that serves as a strict reference for idiomatic JAX styling and test case generation. Second, instead of depending on the LLM to deduce mathematical outputs, we run the source PyTorch modules to get their actual dynamic tensor states. This creates an unchangeable execution oracle. We then use an autonomous agentic loop to synthesize tests based on the oracle data. The test cases are executed repeatedly, and the traceback is sent back to the LLM for self-correction. Ablations show that combining ICL references with oracle grounding and self-debugging greatly outperforms pure instructional and basic agentic baselines. This improvement does not add an excessive computational overhead. Our lightweight pipeline achieves $91\%$ numerical equivalence (compared to baseline: 9\%, instruction + self-debugging: 27\%) on  neural modules, providing a highly reliable, scalable blueprint for cross-framework migration. This has been validated across several state-of-the-art models including SAM (segment anything), T5, Code Whisper amongst others showing high numerical equivalency. 
\end{abstract}

\begin{document}

\maketitle

\section{Introduction}

PyTorch and Jax are popular deep learning frameworks, but have some fundamental differences such as PyTorch's eager-execution style against Jax's JIT (just in time) compilation leveraging XLA (Accelerated linear algebra). This has created a split in modeling ecosystems with unique advantages to each of the ecosystems. Migration of code from one framework to another is often daunting, creating increasing silos over time. This migration is mainly driven by cost and unique advantages to each accelerator-based ecosystem. Google’s Tensor Processing Units (TPUs) offer competitive computation per dollar and highly optimized matrix multiplication speeds.  However, moving from PyTorch code to TPU-optimized JAX code creates a significant engineering challenge. Developers must convert object-oriented, mutable states into JAX's functional, stateless setup using \texttt{flax.linen}. They also need to replace standard Python control flow with compiler-friendly structures like \texttt{jax.lax.scan}. Additionally, developers must address default discrepancies that affect numerical equivalence, such as different tensor layouts (PyTorch uses NCHW while JAX uses NHWC), default precisions (PyTorch uses fp32 while JAX uses bfloat16), and fundamentally different paddings in attention layers.

Modern Large Language Models (LLMs) have shown remarkable proficiency in generating general-purpose code \citep{roziere2023code} including code migrations~\citep{nikolov2026multi}. However, even with improvements in reasoning approaches like Chain of Thought \citep{wei2022chain} and Self-Consistency \citep{wang2022self}, LLMs still struggle with strict framework translation. We believe that LLMs cannot natively handle complex, high-dimensional tensor arithmetic. Without a reliable execution-based verification loop, LLMs tend to hallucinate incorrect mathematical and semantic judgements. To overcome this, we propose a fully automated, agentic pipeline specifically designed for PyTorch-to-JAX translation. We view framework translation as a grounded generation problem. By giving the LLM syntactic anchors for formatting and dynamic execution anchors to define the absolute mathematical ground truth, we create a self-correction loop that ensures that its output aligns with the target system.

The main contributions of this paper are threefold: 
\begin{itemize} 
\item \textbf{Structural Anchoring via In-Context Learning:} We demonstrate that giving a language model a small, carefully chosen set of reference translations and tests for ICL effectively anchors the generation process and lowers the risks of hallucination.
\item \textbf{The Dynamic Execution Oracle:} We introduce a method to work around the arithmetic limitations of LLMs. By running the source PyTorch modules and saving the dynamic computational state to a fixed "Oracle," we align with the shift toward execution-based code evaluation \citep{chen2021evaluating} to ensure mathematical fidelity.
\item \textbf{Ablation Guidance for Agentic Workflows:} We conduct ablation studies that isolate the effects of instructions, ICL, and execution-conditioned self-debugging. By measuring the impact of each part, we offer practical guidance for developing future agentic code-translation systems. \end{itemize}

\begin{figure*}
    \centering
    \includegraphics[width=\linewidth]{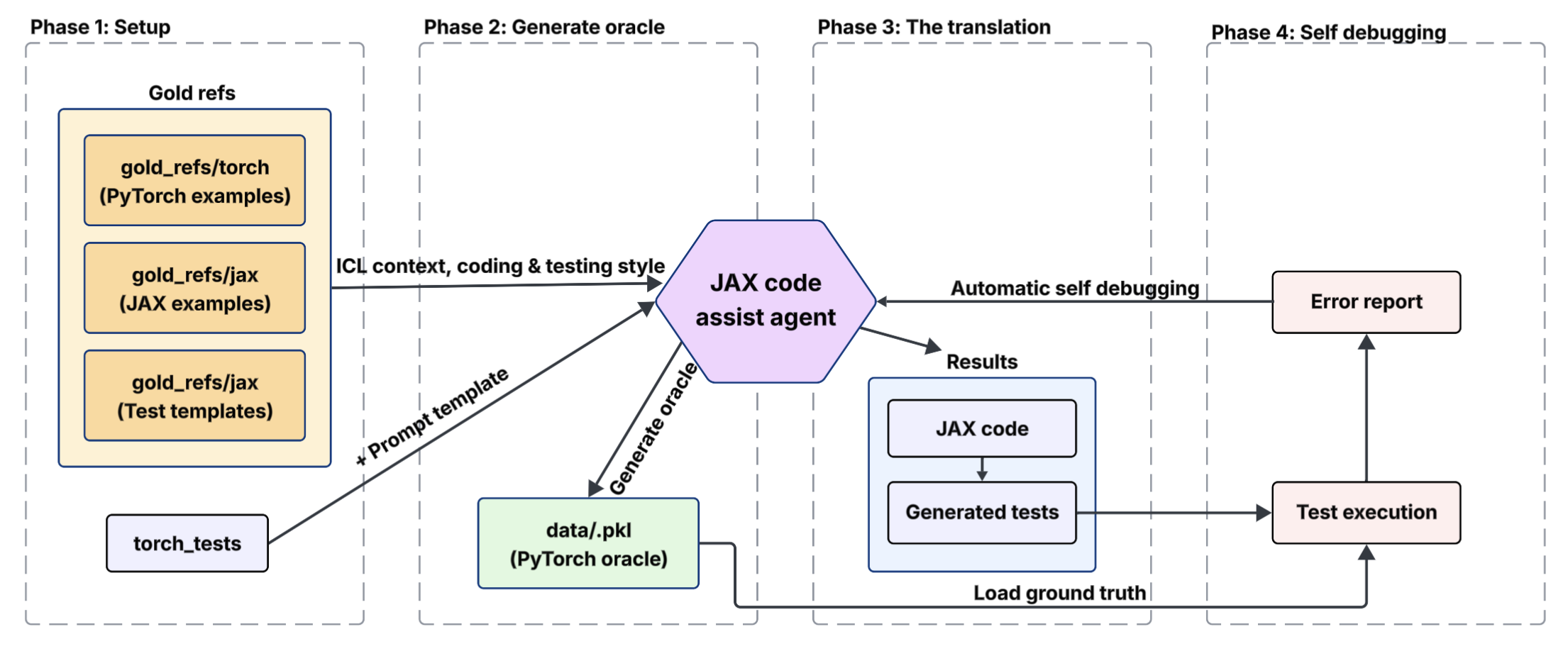}
    \caption{Block diagram of the workflow highlighting the four phase process, including the set-up, generating oracle, the core translation phase and the final self-debugging phase to help the agent autonously self-correct.}
    \label{fig:diagram}
\end{figure*}

\section{Methods}
\label{sec:methods}
We suggest a step-by-step framework for translating PyTorch to JAX. This framework has two main parts: anchoring through In-Context Learning and refinement using feedback from an oracle-driven self-debugging process. To implement this, we divided the workflow into four automated phases, followed by a manual check that provides the final accuracy number we report. Figure~\ref{fig:diagram} represents the general workflow of our approach. In the following, we describe each phase in detail.

\subsection{Phase 1: Anchoring Translation via In-Context Learning (ICL)}
The main challenge in translating PyTorch to JAX automatically is to move from mutable, object-oriented states to a functional, stateless design. We first tried a Retrieval-Augmented Generation (RAG) pipeline referencing Google’s MaxText repository. This approach created significant problems. The fragments retrieved introduced complex issues related to distributed computing, such as hardware sharding. These distractions worsened the model's API hallucinations, illustrating the "Lost in the Middle" phenomenon (\citep{liu2024lost}). To reduce this repository-specific noise, we used a tightly controlled In-Context Learning (ICL) method. Research shows that large models can apply ICL to overwrite pre-trained patterns and compact demonstrations into task vectors (\citep{hendel2023context}) that act as structural anchors for clear syntactic formatting (\citep{min2022rethinking}). Instead of a broad retrieval, we found that providing a guided boundary and execution feedback greatly boosts accuracy. Thus, our context is limited to a small but dense set of references with rich and diverse reasoning. This includes the source PyTorch module, the idiomatic JAX translation, and the relevant test cases. This setup provides structural guidance to anchor the functional framework without overwhelming the context window.

\subsection{Phase 2: Dynamic Execution and the PyTorch Oracle}
To ensure code functional equivalence, our framework treats the original PyTorch implementation as the unquestioned ground truth. We assume the source module is mathematically correct and aim to mirror its functionality. Relying on an LLM to read code and predict high-dimensional tensor arithmetic can cause critical mistakes. Models might produce incorrect results for complex mathematical operations and intermediate shapes (\citep{dziri2023faith}). Such mistakes can be very expensive to debug, and might just trade-off code migration effort with debugging effort. Therefore, it is critical that agents avoid or catch such critical mistakes autonomously.

Towards this, instead of heuristics , we direct the agent to create a profiling script for dynamic execution. This script compiles the PyTorch module, passes random input tensors through the forward pass, and extracts the computational state. This state includes the initialized weights ($state_{dict}$), input tensors, and output activations. We serialize this state into an unchangeable artifact, which we call our "Oracle." Importantly, running the code on the PyTorch backend reveals behaviors that static analysis misses. These behaviors include silent padding methods and shape changes during execution. By basing our evaluation on this immutable Oracle, we avoid the limitations of the LLM's arithmetic. This ensures that the downstream JAX translation is validated against real computation instead of theoretical estimations.

\subsection{Phase 3: Agentic Translation and Oracle-Conditioned Test Generation
}
Grounded by the structural anchors from Phase 1, the agent translates the source PyTorch modules into standard flax.linen abstractions. Constrained by the selected ICL examples, the model carefully navigates the shift in framework, producing functional forward passes. However, simply generating syntactically correct code is not enough; verifying semantic equivalence is still the main challenge. To automate this verification, we direct the agent to create custom test harnesses based directly on the Phase 2 Oracle and Phase 1 ICL examples. 

\begin{table*}[htbp]
\centering
\caption{Ablation Study Results Across Level 1 and Level 2 data comparing completeness, shape equivalence and numerical equivalence.}
\label{tab:ablation_results}
\resizebox{\textwidth}{!}{%
\begin{tabular}{lccc ccc ccc}
\hline
\textbf{Configuration} & \textbf{Instruction} & \textbf{ICL} & \textbf{Iterative} & \multicolumn{3}{c}{\textbf{Level 1 (Mathematical Ops)}} & \multicolumn{3}{c}{\textbf{Level 2 (Neural Modules)}} \\
\textbf{Configuration} & \textbf{Prompts} & \textbf{Anchor} & \textbf{Debugging} & \multicolumn{3}{c}{} & \multicolumn{3}{c}{}\\
\cline{5-10}
 & & & & \textbf{Comp.} & \textbf{Shape} & \textbf{Num. Eq.} & \textbf{Comp.} & \textbf{Shape} & \textbf{Num. Eq.} \\
\hline
Baseline Only & $\times$ & $\times$ & $\times$ & 44\% & 44\% & 44\% & 18\% & 18\% & 9\% \\
Instruction Only & \checkmark & $\times$ & $\times$ & 67\% & 67\% & 44\% & 36\% & 36\% & 18\% \\
Instruction + Self-Debugging & \checkmark & $\times$ & \checkmark & 89\% & 89\% & 89\% & 73\% & 55\% & 27\% \\
\textbf{Full Pipeline (Ours)} & \checkmark & \checkmark & \checkmark & \textbf{100\%} & \textbf{100\%} & \textbf{100\%} & \textbf{91\%} & \textbf{91\%} & \textbf{91\%} \\
\hline
\end{tabular}%
}
\vspace{0.1cm}
\end{table*}
The evaluation follows a hierarchical three-stage process. First, the script checks if the code can compile, acting as a strict pass/fail test. It ensures the JAX module initializes and traces correctly without runtime errors. Second, it checks for shape correctness (pass/fail), confirming dimensional consistency and addressing framework layout changes (for example, mapping PyTorch’s NCHW to JAX’s NHWC). Finally, the script programmatically deserializes the .pkl files, maps the PyTorch weights into the JAX model, and checks for numerical equivalence. For this last stage, we set a strict absolute error limit of $10^{-7}$ using numpy.allclose. This specific tolerance separates structural translation errors from the natural limits of standard 32-bit floating-point numbers (Higham, 2002), resulting in a reliable testing harness we refer to as the "Silver" test cases.

\subsection{Phase 4: Iterative Self-Debugging and Silver Reference Validation}
Initial zero-shot translations sometimes fail due to JIT compilation errors, unhandled PRNG routing, or strict numpy.allclose mismatches. To prevent these issues, we embed structural instructions, such as PRNG key splitting constraints, into the generation prompt. However, static instructions alone do not catch every edge case. An iterative self-debugging loop can effectively manage the challenges of cross-framework migration. When a generated module fails the Oracle-conditioned test, the execution feedback, including standard error, stack traces, and precise numerical differences, are fed directly to the agent's context. With this objective feedback and guided by the ICL templates, the LLM addresses its initial errors and refines the code from a broken state into a reliable, mathematically verified JAX module.

While this automated test generation and debugging loop drives our system's capability, we must protect our final metrics from reward hacking, such as the agent creating trivial tests to pass. To validate our approach scientifically, we separate the debugging capabilities from the final evaluation benchmark. The translation accuracies presented in this work are measured only against a manual evaluation suite. Furthermore, the modules used for ICL do not appear in this evaluation set. This strict separation ensures our reported metrics reflect true, uncontaminated generalization, demonstrating that our grounded, agent-driven pipeline effectively addresses the framework translation bottleneck.

\section{Benchmark}
To quantify the impact of our architectural anchors, we evaluated four pipeline configurations on a holdout set of human-crafted tests. To rigorously test scaling complexity, we categorize our evaluation into three difficulty tiers. \textbf{Level 1} comprises 9 fundamental mathematical and training operations (\textit{data mixup, early stopping, gradient clipping, positional encoding, moving average, and dice, focal, label smoothing, and tversky losses}). \textbf{Level 2} evaluates 11 architectural abstractions and neural modules (\textit{MLPs, VAEs, GANs, GRUs, LSTMs, attention mechanisms, transformer encoders, temporal convolutional units, graph convolutions, and ResNet blocks}). \textbf{Level 3} evaluates migration of an entire repository (given as a Github or HuggingFace link) to JAX. Since it is complex to evaluate Level 3 automatically, we leverage human experts to evaluate for this task. Ten repositories were included for level 3 which include Code-Whisper, Multi-modal transformers (MMTF), simple-moe, time-series forecasting model, FinGPT model (finance GPT), StyleGAN, T5 transformer, two-tower recommendation model, SAM (segment anything) model, detr-resnet50. \par

For level1 and level 2, we measure structural validity (compilation success), layout consistency (shape correctness), and mathematical fidelity. Given a source PyTorch module $f_{pt}$, a generated JAX module $f_{jax}$, identically initialized and mapped weights ($W_{pt}$, $W_{jax}$), and an input tensor $X \sim \mathcal{D}$, numerical equivalence is formally defined as:$$\max \left| f_{pt}(X, W_{pt}) - f_{jax}(X, W_{jax}) \right| < \epsilon$$where $\epsilon = 1 \times 10^{-7}$ accounts for standard floating-point disparities between backend compilers.\par
For level 3, since the evaluations are by human evaluators, we only report the results of the full pipeline (ours). The completeness of modules/files converted and the numerical equivalence are reported. In addition, they also report an average readability score on a Likert scale of 1-5 (averaged across all the files in a repository). Since readability is subjective, we only report the scores without claims of broader implications.
\section{Results}
\label{sec:ablations}

 Table~\ref{tab:ablation_results} summarizes results comparing our pipeline to the baselines, instructions, and the self-debugging loops. Although self debugging significantly improved performance over the instruction only prompts, there was still a wide gap for usability of the code. Although self-debugging and detailed instructions helped improve average numerical equivalence to between 27\% and 89\% across levels 1 and 2 (module and layer level). The shape matching reached 55\% on Level 2 with the self-debugging loop. However, with our full pipeline, the shape and numerical equivalence reached 91\% on level 2 and 100\% on level 1 respectively.  \par
 
 Table~\ref{tab:level3_results} summarizes results for the level 3 repository migration dataset. The inputs are full code repository (level 3), and human experts used the code migration tool independently of developers, were allowed to make minor edits (between 1-2) to guide the agent but have it run mostly autonomously. In general, the agent was able to complete the task almost always. Numerical equivalency was 100\% for four of the repositories. However, for two of them, it fell below 85\%, which were both from facebook research and were foundational and expansive models. \par

\textbf{Taxonomy of Translation Failures:} Across the evaluation levels, zero-shot and static-instruction baselines showed three main types of failures. First, Syntactic Hallucinations:  the LLM often did not generate the necessary wrapper classes, causing frequent \texttt{NameError}s or missing \texttt{init} attributes. Second, API Signature Mismatches: the model mistakenly applied PyTorch-specific kwargs, such as passing \texttt{train=True} directly into a \texttt{\_\_call\_\_} method). Finally, State Mapping Discrepancies: even when modules compiled, even when modules compiled, complicated structural translations led to \texttt{KeyError}s during the precise weight-transfer phase like mismatched \texttt{Dense} layers in attention blocks. This highlighted the need for a self debugging loop.

\begin{scriptsize}
\begin{table}[h]
\centering
\caption{Level3 Human Evaluation across 10 github repositories}
\label{tab:level3_results}
\begin{footnotesize}
\begin{tabular}{ccccc}
\hline
\textbf{Repository} & \textbf{\footnotesize{Github link (prefix: https://github.com/)}} & Completeness & Num. Eq. & Readability \\
\hline
Code Whisper & \footnotesize{openai/whisper/tree/main/whisper} & 100\% & 100\% & 4 \\
\footnotesize{Multimodal transformer} & \footnotesize{yaohungt/Multimodal-Transformer} & 100\% & 100\% &  4.3 \\
MOE & \footnotesize{peytontolbert/simple-moe} & 100\% & 93.3\% &  4.5 \\
Time-series forecasting & \footnotesize{jinglescode/time-series-forecasting-pytorch} & 100\% & 96.5\% &  4.8 \\
FinGPT & \footnotesize{AI4Finance-Foundation/FinGPT} & 100\% & 86\% &  4.9 \\
StyleGAN2 & \footnotesize{NVlabs/stylegan2-ada-pytorch} & 100\% & 96\%  & 4.4 \\
T5 & \footnotesize{google-research/text-to-text-transfer-transformer} & 100\% & 100\%  & 3.75 \\
Two-tower & \footnotesize{gauravchak/two\_tower\_models} & 100\% & 100\%  & 4.9 \\
Segment Anything (SAM2) & \footnotesize{facebookresearch/segment-anything} & 86\% & 80\%  & 4.8 \\
DETR & \footnotesize{facebookresearch/detr} & 100\% & 57\%  & 3.8 \\
\hline
\end{tabular}%

\vspace{0.1cm}
\end{footnotesize}
\end{table}
\end{scriptsize}
\begin{itemize}
    \item \textbf{Baseline:} Simple prompting predictably fails due to the previously mentioned syntactic hallucinations. The LLM creates PyTorch-like mutable states in JAX syntax, resulting in immediate compilation failures and almost no equivalence. 
    \item \textbf{Instruction Only:} Providing detailed translation prompts, such as clear rules on PRNG routing and \texttt{flax.linen} requirements, helps reduce basic \texttt{NameError}s and slightly improves compilation rates. However, relying only on written instructions does not guarantee all the necessary structural understanding. This approach works partially for Level 1 tasks, which are mainly stateless and need only localized tensor manipulations, but it was not that helpful for Level 2 modules. The model has trouble with complex state management, hierarchical initialization, and hidden dimensional shifts needed for neural architectures. As a result, mathematical accuracy stalls at 44\% for Level 1 and only 18\% for Level 2.
    \item \textbf{Detailed Prompt Instruction + Self Debugging:}  Adding a compiler-driven feedback loop without the Oracle creates a false sense of success. The agent fixes syntactical errors, increasing the compilation rate. However, without a clear mathematical truth, the LLM resorts to reward hacking to achieve a compilation pass, which leaves numerical equivalence disproportionately low, especially on complex Level 2 architectures.
    \item \textbf{Full Pipeline (Ours):} 
    The complete framework shows that structural anchoring must be combined with empirical truth. By requiring the generated test harnesses to physically load the serialized Oracle `.pkl` data, the debugging loop must iterate on its architecture and weight-mapping logic until it meets the strict mathematical condition ($< \epsilon$). This leads to the highest overall fidelity, achieving 100\% fidelity on Level 1 and 91\% on Level 2, demonstrating the need for execution-backed constraints.
\end{itemize}

\section{Conclusion}

Translating deep learning research from PyTorch to optimized JAX environments remains a persistent challenge. We demonstrate that while LLMs natively struggle with this shift, they are highly capable code generators when given clear rules and good references. By combining structural In-Context Learning anchors with dynamic feedback from an oracle, our fully automated system can handle the complexities of functional state management and hidden numerical issues. While our evaluation shows high fidelity on independent modules, we acknowledge two main limitations: our method has not yet been expanded to fully automatic repository level migrations, which will need complete structural dependency management, and we have not systematically explored how changing the specific examples used in the ICL affects results. Still, this grounded, agentic approach eliminates the need for manual test and offers a scalable blueprint for high-fidelity cross-framework migration.

\bibliographystyle{plainnat}
\bibliography{refs}

@article{liu2024lost,
  title={Lost in the middle: How language models use long contexts},
  author={Liu, Nelson F and Lin, Kevin and Hewitt, John and Paranjape, Ashwin and Bevilacqua, Michele and Petroni, Fabio and Liang, Percy},
  journal={Transactions of the association for computational linguistics},
  volume={12},
  pages={157--173},
  year={2024}
}

@inproceedings{hendel2023context,
  title={In-context learning creates task vectors},
  author={Hendel, Roee and Geva, Mor and Globerson, Amir},
  booktitle={Findings of the Association for Computational Linguistics: EMNLP 2023},
  pages={9318--9333},
  year={2023}
}

@inproceedings{min2022rethinking,
  title={Rethinking the Role of Demonstrations: What Makes In-Context Learning Work?},
  author={Min, Sewon and Lyu, Xinxi and Holtzman, Ari and Artetxe, Mikel and Lewis, Mike and Hajishirzi, Hannaneh and Zettlemoyer, Luke},
  booktitle={Proceedings of the 2022 Conference on Empirical Methods in Natural Language Processing},
  pages={11048--11064},
  year={2022}
}

@inproceedings{dziri2023faith,
  title={Faith and Fate: Limits of Transformers on Compositionality},
  author={Dziri, Nouha and Lu, Ximing and Sclar, Melanie and Li, Xiang Lorraine and Li, Jian and Lin, Bill Yuchen and West, Peter and Bhagavatula, Chandra and Le Bras, Ronan and Hwang, Jena D and others},
  booktitle={Advances in Neural Information Processing Systems},
  volume={36},
  year={2023}
}

@article{roziere2023code,
  title={Code llama: Open foundation models for code},
  author={Roziere, Baptiste and Gehring, Jonas and Gloeckle, Fabian and Sootla, Sten and Gat, Itai and Tan, Xiaoqing Ellen and Adi, Yossi and Liu, Jingyu and Sauvestre, Romain and Remez, Tal and others},
  journal={arXiv preprint arXiv:2308.12950},
  year={2023}
}

@article{wei2022chain,
  title={Chain-of-thought prompting elicits reasoning in large language models},
  author={Wei, Jason and Wang, Xuezhi and Schuurmans, Dale and Bosma, Maarten and Xia, Fei and Chi, Ed and Le, Quoc V and Zhou, Denny and others},
  journal={Advances in neural information processing systems},
  volume={35},
  pages={24824--24837},
  year={2022}
}

@article{wang2022self,
  title={Self-consistency improves chain of thought reasoning in language models},
  author={Wang, Xuezhi and Wei, Jason and Schuurmans, Dale and Le, Quoc and Chi, Ed and Narang, Sharan and Chowdhery, Aakanksha and Zhou, Denny},
  journal={arXiv preprint arXiv:2203.11171},
  year={2022}
}

@article{chen2021evaluating,
  title={Evaluating large language models trained on code},
  author={Chen, Mark and Tworek, Jerry and Jun, Heewoo and Yuan, Qiming and Pinto, Henrique Ponde De Oliveira and Kaplan, Jared and Edwards, Harri and Burda, Yuri and Joseph, Nicholas and Brockman, Greg and others},
  journal={arXiv preprint arXiv:2107.03374},
  year={2021}
}

@article{nikolov2026multi,
  title={A Multi-agent AI System for Deep Learning Model Migration from TensorFlow to JAX},
  author={Nikolov, Stoyan and Konrad, Bernhard and Gronbach, Moritz and Kumar, Niket and Yan, Ann and Singh, Varun and Liang, Yaning and Ranganathan, Parthasarathy},
  journal={arXiv preprint arXiv:2603.27296},
  year={2026}
}

\appendix

\end{document}